\documentclass[runningheads]{llncs}
\usepackage[T1]{fontenc}
\usepackage{hyperref}
\usepackage{graphicx}
\usepackage{amsmath}
\usepackage{wrapfig}
\usepackage{booktabs}

\begin{document}
\title{BiasScanner: Automatic Detection and Classification of News Bias to Strengthen Democracy\thanks{The authors gratefully acknowledge the funding provided by the Free State of Bavaria under its ``Hitech Agenda Bavaria''. All views are the authors' and do not necessarily reflect the views of any funders or affiliated institutions. We thank Michael Reiche for feedback and discussions.}}
\titlerunning{BiasScanner}
%

\author{Tim Menzner\inst{1} \and Jochen L.~Leidner\inst{1,2}} 
\authorrunning{T.~Menzner and J.~L.~Leidner}


\institute{Coburg University of Applied Sciences, Coburg, DE \and University of Sheffield, Sheffield, UK\\(Contact: \email{leidner@acm.org})}
\maketitle              
\begin{abstract}
  The increasing consumption of news online in the 21st century
  coincided with increased publication of disinformation, biased reporting, hate speech and other unwanted Web content. \\
  We describe \emph{BiasScanner}, an application
  that aims to strengthen democracy by supporting news consumers with scrutinizing news articles they are reading online. BiasScanner contains a server-side
  pre-trained large language model to identify biased sentences of news articles and a front-end Web browser plug-in. 
  At the time of writing, BiasScanner can identify and
  classify more than two dozen types of media bias at the sentence level,
  making it the most fine-grained model and only deployed application (automatic system in use) of its kind.
  It was implemented in a light-weight and privacy-respecting
  manner, and in addition to highlighting likely biased
  sentence it also provides explanations for each
  classification decision as well as a summary analysis for each news article. 
  
  While prior research  has addressed news bias detection, we are not aware of any work that resulted in a deployed browser plug-in (c.f. also
  \url{biasscanner.org} for a Web demo).
  
   \and information access systems
\end{abstract}


\section{Introduction} \label{sec:intro}

Democracy faces an existential threat when most citizens get their news from online platforms focused on controversy rather than balanced reporting. Such behavior increases advertising revenue, contributing to media bias and the spread of fake news \cite{Lee-Solomon:1990,Groseclose-Milyo:2005:QuartJEcon,Sloan-Mackay:2007,Groeling:2016:AnnuRevPolitSci,Vosoughi-Roy-Aral:2018:Sci}.

To combat this trend, \textbf{we introduced \emph{BiasScanner}, a practical tool to help readers assess online news regarding instances of biased reporting}, which we describe here. It highlights biased sentences, offers detailed analysis reports, and assigns bias scores. Users can also donate bias reports for research. BiasScanner makes use of advanced neural transformer models, such as OpenAI's GPT 3.5 for efficient and effective bias detection. We prioritize user privacy by not storing personal(ly identifiable) information or news stories without explicit consent.


\section{Related Work} \label{sec:related-work}

\textbf{Foundational Language Models.}
The neural transformer model was first described in 
\cite{Vaswani-etal:2017:NeurIPS}. Google BERT \cite{Devlin-etal:2019:NAACL} and 
OpenAI's GPT-3 \cite{Brown-etal:2020:NeurIPS} GPT-4 (and their application ChatGPT \cite{Roumeliotis-Tselikas:2023:FutureInternet}) and Meta's Llama \cite{Touvron-etal:2023:ArXiv} have been early foundational
models that have introduced a paradigm shift in NLP by demonstrating
how large, pre-trained language models can dramatically reduce
the development time of NLP systems by using large quantities of
un-annotated text to train general-purpose ``foundational'' models. 

\textbf{News Bias.} Groeling \cite{Groeling:2016:AnnuRevPolitSci} presents
a survey of the literature covering partisan
bias. Conrad et al. \cite{Conboy:2007} focused on content mining to measure credibility of authors on the web. The topic of bias in mass media was dealt with in detail by \cite{Lee-Solomon:1990} and \cite{Sloan-Mackay:2007}. Hamborg et al. \cite{Hamborg-etal:2020:JCDL} provided an interdisciplinary literature review to suggest methods how bias could be bias detection could be automated. \\
\textbf{Bias Detection.}
Media bias datasets with different focus where released by \cite{Arapakis-etal:2016:ACL},\cite{Horne-etal:2018:ArXiv}, and \cite{Spinde-etal:2021:IPM,Spinde-Hamborg-Gipp:2020:ECMLPKDD}
After early pioneering work on bias from economics \cite{Groseclose-Milyo:2005:QuartJEcon}, Arapakis et al. \cite{Arapakis-etal:2016:ACL} labeled 561 articles along 14 quality dimensions including subjectivity.
Horne et al. \cite{Horne-etal:2018:ArXiv} released a larger dataset annotated for political partisanship bias, but without grouping articles by event, which makes apples-to-apples comparison harder;
Chen et al. \cite{Chen-etal:2018:ICNLG} addressed this issue by resorting to another corpus sampled from the website \url{allsides.com}, which includes human labels by U.S. political orientation (on the ordinal scale $\{LL, L, C, R, RR\}$); they also present an ML model to flip the orientation to the oppositite one.
Yano, Resnik and Smith 
\cite{Yano-Resnik-Smith:2010:NAACLWS} also on
the liberal-conservative axis, manually
annotating sentence-level partisanship bias.\\
MBIB, the first media bias identification benchmark, was introduced by Wessel et al. \cite{Wessel-etal:2023:SIGIR}, who evaluated Transformer techniques on detecting nine different types of bias across 22 selected datasets. 
Baumer \textit{et al.} focused on detecting framing language. \cite{Baumer-etal:2015:NAACL}
Chen et al. \cite{Chen-etal:2020:FindEMNLP} demonstrated that incorporating second-order information, such as the probability distributions of the frequency, positions, and sequential order of sentence-level bias, can enhance the effectiveness of article-level bias detection, especially in cases where relying solely on individual words or sentences is insufficient.
Spinde et al. published a dataset containing biased sentences and evaluated detection techniques on it. \\
\cite{Spinde-etal:2021:IPM,Spinde-Hamborg-Gipp:2020:ECMLPKDD}

\textbf{Web Apps \& Mobile Apps.}
Hamborg et al. \cite{Hamborg-etal:2020:JCDL}
presented \emph{Newsanalyze}, a system that
highlights sentiment target entities colored by
polarity. In contrast, we perform sentence
classification targeting bias and sub-type of
bias (sentiment $=$ affective state $\neq$ bias (Although there can and typically is a connection, bias is more general e.g. under-reporting is not sentiment-related at all). Da San Martino et al \cite{da-san-martino-etal-2020-prta} developed Prta, a tool highlighting propaganda techniques in news articles. While propaganda and news bias are related (as visible in the overlap of propaganda techniques and bias types), new bias is a broader phenomena, also including unintentional subjective reporting.\\
\textbf{Other Related Work.} Conrad, Leidner and Schilder characterize signals for credibility in the context of credibility for professionals \cite{Conrad-Leidner-Schilder:2008:WICOWWS}. Bhuiyan et al. \cite{Bhuiyan-etal:2020:HCI} compare crowdsourced and expert assessment criteria for credibility on statements about climate change. Allen and co-workers
\cite{Allen-etal:2021:SciAdv} studied the
Ghanem et al. \cite{Ghanem-etal:2021:EACL} analyze an interesting way to
distinguish between real/credible news and
fake news by looking at the distribution of
affective words within the document.

To the best of our knowledge, BiasScanner is the first system for news bias detection and bias sub-type classification based on a neural transformer architecture published in the scientific literature and deployed/release to the general public as a free browser plug-in.



\sloppypar{}\section{System} \label{sec:system}

This section describes \textbf{BiasScanner}, our system, which is also deployed on the World Wide Web at \url{https://biasscanner.org}. This address also
contains a separate Web demo where users can experiment
with our model before installing the Web browser plug-in.

\subsection{Architecture} \label{subsec:architecture}

\textbf{Architecture.} 
We designed BiasScanner with ease and convenience of use and
 respect for the user's privacy in mind. A frond-end application
 deals with the user interface and communicates with our server,
 which provides a bias classification service, and which shields
 the originating IP address of the user when invoking OpenAI GPT -- current model: a gpt-3.5-turbo-16k fine-tuned on articles constructed from the BABE dataset \cite{Spinde2021f} with information about bias type and strength added using GPT-4 -- via a REST API on a US server, but without any transfer of PII data. Our server
 layer also deals with payment authentication for the transformer
 model use to hide this aspect from users, as we believe dealing with cumbersome API keys would
 exclude some users. The nature of our architecture also permits easy switching of the model working behind the scenes (we are considering switching to an Open Source Model long-term) without disruption for users.  

\begin{figure}
    \centering
    \includegraphics[width=.5\textwidth]{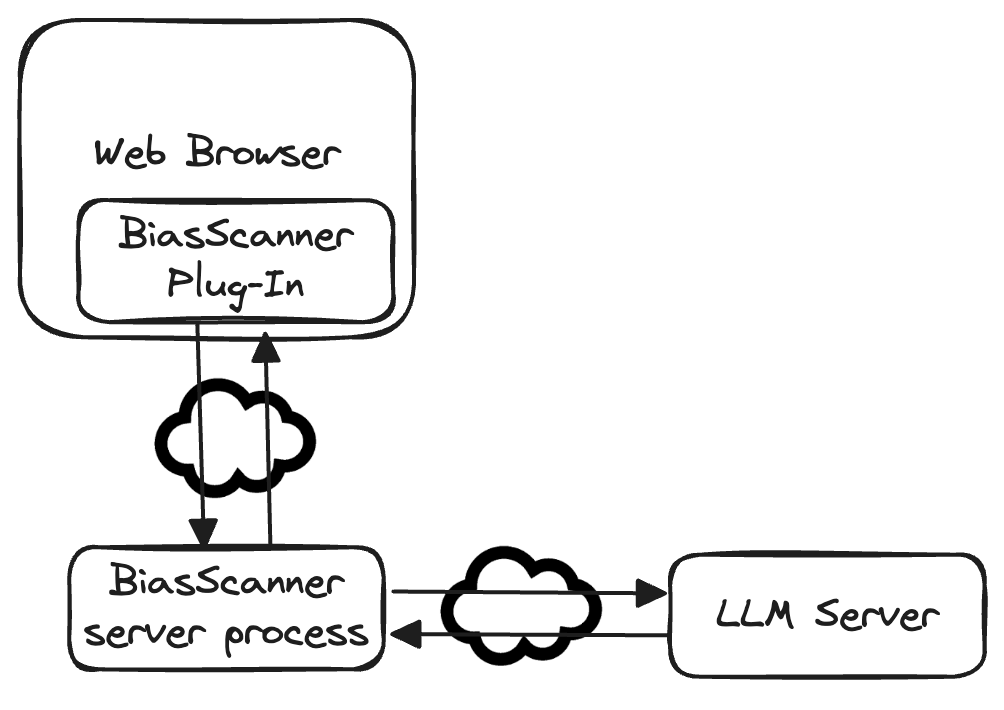}
    \caption{BiasScanner System Architecture}
    \label{fig:system-architecture}
\end{figure}

We designed BiasScanner in a user-friendly way and with  privacy protection in mind. The overall architecture is shown in Figure \ref{fig:system-architecture}. Our front-end application, a Web browser plug-in, handles the user interface and connects to our server, which offers bias classification, currently in turn calling OpenAI via a US-based REST API as its large language model (LLM) server, with user IP address protection and no PII data transfer. Additionally, our server manages payment authentication to simplify the user experience; we aim to avoid the hassle of dealing with API keys, ensuring inclusiveness
for all users.\\

\textbf{Implementation.} We implemented BiasScanner as a Web Application on our site, where users can type or copy in text to get an analysis and as browser plug-in for Firefox, Chrome/Chromium and other browsers using JavaScript. When using the plug-in, the relevant article text is extracted from the HTML of a web page by utilizing Mozilla's readability library, which also serves as base of the Firefox reader view \cite{mozilla_readability}.

\subsection{User Interface} \label{subsec:gui}

\textbf{User Interface.} 

\begin{wrapfigure}{l}{0.65\textwidth}
        \includegraphics[width=0.65\textwidth]{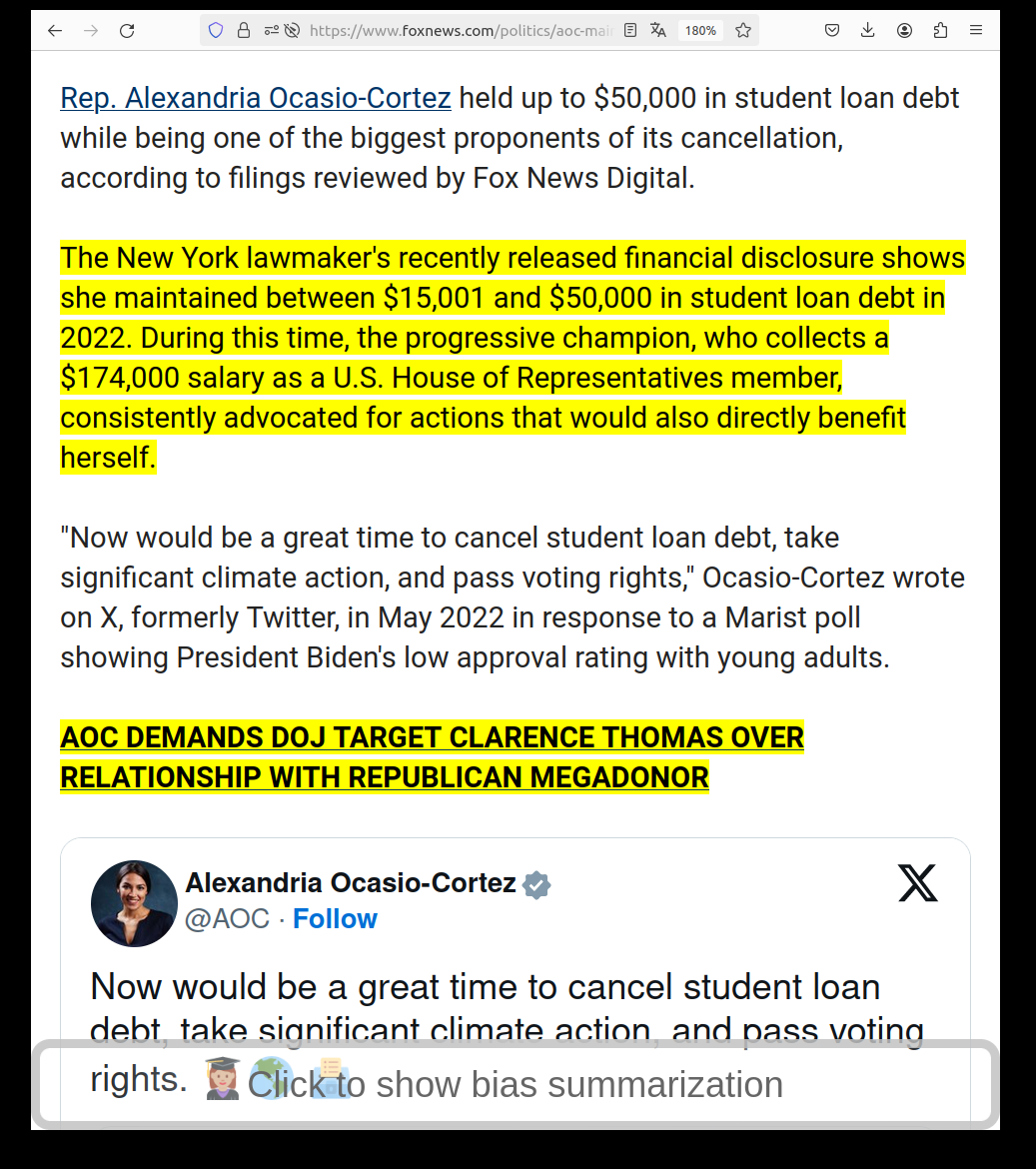}
        \caption{In-line highlighting of biased sentences}\label{fig:gui}
\end{wrapfigure}

Figure \ref{fig:gui} shows the graphical user interface of BiasScanner (Web plug-in version).


The prompt used for instructing the language model
was developed iteratively and aims to provide
consistent and high-quality output by considering
best practices, like a clear definition of every searched-for bias type and by including an example for the desired JSON output format. 
The answer given by the model is then post-processed and filtered to prevent potential errors before being used to highlight biased sentences directly on the site. A more detailed report including the type of bias,  a short explanation and a score indicating the strength of the bias, is also available for the user to view. This bias report concludes by providing a general assessment of the article's bias(es). 

It calculates a score by normalizing the sum of two components: the ratio of biased sentences to total sentences in the article and the average bias score across all biased sentences in the article.
The prompt for instructing the language model was developed in several iterations to ensure consistent and high-quality output. It includes a clear definition of each searched-for bias type and an example for the desired JSON output format. The model's response is post-processed and filtered to prevent errors before highlighting biased sentences on the site. Users can access a detailed report that includes bias type, explanation, and a bias strength score. This report also provides a general assessment of the article's bias, and a overall score, calculated by normalizing the ratio of biased sentences to total sentences and the average bias score across all biased sentences. \\

\textbf{Currently Supported Types of Bias.}
In general, we define media bias as the tendency to, consciously or unconsciously,
report a news story in a way that supports a pre-existing narrative instead of
providing unprejudiced coverage of an issue.
Our implementation explicitly searches for 27 different types of Bias, namely Ad Hominem Bias, Ambiguous Attribution Bias,  Anecdotal Evidence Bias, 
Causal Misunderstanding Bias, Cherry Picking Bias, 
Circular Reasoning Bias, Discriminatory Bias, Emotional Sensationalism Bias, External
Validation Bias, False Balance Bias, False
Dichotomy Bias, Faulty Analogy Bias, Generalization Bias, Insinuative Questioning Bias, Intergroup
Bias, Mud Praise Bias, Opinionated Bias, 
Political Bias, Projection Bias, Shifting Benchmark Bias, Source Selection Bias, Speculation Bias, Straw Man Bias, Unsubstantiated Claims Bias, Whataboutism Bias and Word Choice Bias:




\begin{figure}
    \hspace{-0.8cm}
            \includegraphics[width=1.1\textwidth]{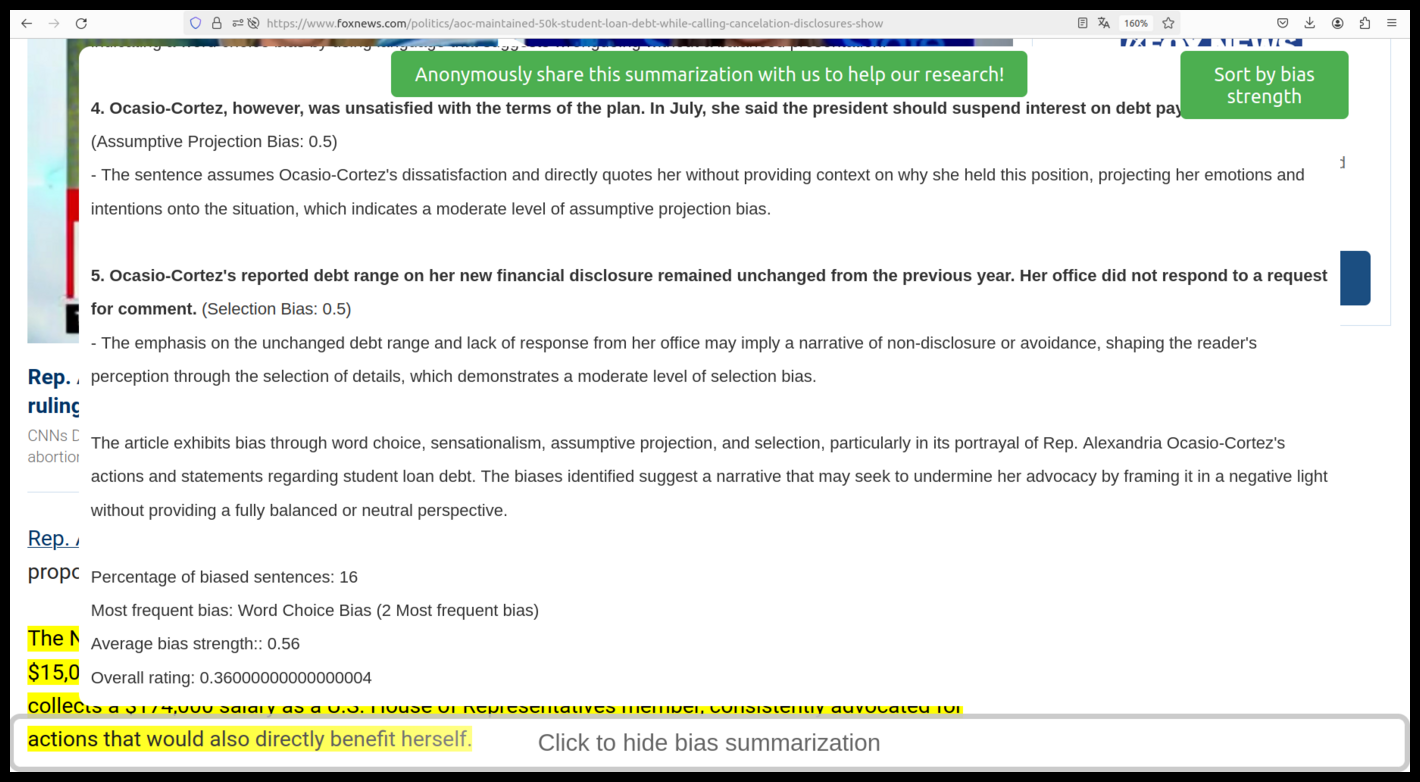}
    \caption{Bias Summary Report}
    \label{fig:enter-label}
\end{figure}



\section{Evaluation} \label{sec:eval}

\textbf{Quantiative Evaluation.} While a detailed evaluation is beyond the
scope of this system paper, we presented detailed
quantitative and qualitative evaluations for the English language in
\cite{Menzner-Leidner:2024:ECIR} and \cite{Menzner-Leidner:2024:NLDB}. Table \ref{table:eval} shows some
quality numbers from  \cite{Menzner-Leidner:2024:NLDB},
which were representative as of June 2024 (for BiasScanner as of release 1.0.0 from July 2024; ongoing development may lead to different scores going forward.)
F1-score is high at 76\%, and our fine-tuned model's quality dominates GPT-4 on all metrics except for precision (73\% versus the latter's 85\%, at the time of writing).\\

\begin{table}
\vspace{-10pt}
\caption{Evaluation Results on the BABE dataset for BiasScanner, GPT-3.5-turbo-1106 with prompt only and GPT-4-turbo-0125. Best results are highlighted in bold.}
\label{table:eval}
\begin{tabular}{lrrrrrrrr}
\toprule
\textbf{Model} & \textbf{TP} & \textbf{FP} & \textbf{FN} & \textbf{TN} & \textbf{F1-Score} & \textbf{Recall} & \textbf{Precision} & \textbf{Accuracy} \\
\midrule

BiasScanner & 576 & 214 & 154 & 524 & \textbf{0.758} & 0.790 & 0.729 & \textbf{0.749} \\
GPT-3.5 (Zero shot) & 384 & 205 & 346 & 533 & 0.582 & 0.526 & 0.651 & 0.624 \\
GPT-4.0 (Zero shot) & 393 & 69 & 337 & 669 & 0.659 & 0.538 & \textbf{0.850} & 0.723 \\
Baseline (Random) & 362 & 374 & 368 & 364 & 0.494 & 0.496 & 0.492 & 0.495 \\

\bottomrule
\end{tabular}
\vspace{-10pt}
\end{table}

\textbf{Qualitative Evaluation.} The achieved quality level
is satisfying for practical use of the browser plug-in; a common error is the mis-classification of neutral reporting
sentences with embedded radical quotes as ``biased''; we
believe embedded quotes ought to be removed before judging
a sentence, which we will address in future work.
We are particularly encouraged by the quality of
our generated explanations, the evaluation of which is
left for future work.\\

\textbf{Beyond English.} 
At the time of writing, BiasScanner can deal with news in
English through our fine-tuned model, and also with other
languages via said model's transfer capabilities; in future
work we want to fine-tune models for additional specific
languages and evaluate them, as well as compare their
performance with our existing model's transfer abilities.





\section{Limitations and Ethical Concerns} \label{sec:limitations}

BiasScanner may not identify all
instance of biases, and while we do not claim
it does, the users may wrongly believe otherwise,
consciously or unconsciously, after getting used to it. It can also not recognize all types of bias: notably,
underreporting bias and other types that need across
across several articles, are beyond its scope, as it only analyzes one individual news story at a time; we leave news coverage comparison for future work. It should also be noted that bias detection
is always, to an extent, a subjective matter. Often a sentence might be considered biased by one person while another considers it to still be objective, therefore no classification will probably ever satisfy everyone at once.  

Our current back-end implementation still depends on an
underlying proprietary foundational model; in future work,
we plan to become independent and port to an open model,
even if this may mean a slight reduction of accuracy, as
this may limit the ability to manipulate the system's
behavior from the outside.


\section{Summary, Conclusions/Limitations and Future Work} \label{sec:conclusion}

We introduced BiasScanner, a new system for enhancing online news consumption by highlighting biased individual sentences in news articles, by offering news story analysis within Web browsers.
We have successfully realized our design goals, including user privacy, rapid implementation and accurate bias classification.

BiasScanner may not identify \emph{all} biases, as to date it focuses on individual news stories and does not compare across articles. 

To date, BiasScanner has mainly been tested with English articles, introducing a development bias. Sending plain text to a server for security is required, but it is done anonymously.
The system has been released as experimental browser extension available free of charge for Firefox trough the Mozilla plug-in marketplace\cite{firefox:biasscanner2024} (Available on Desktop and Android). Future Releases for Chrome and Safari are planned. It can also be installed from GitHub \cite{BiasScannerRepo}.

We are also already using BiasScanner
in the classroom for the teaching of critical reading and engaging students with the topic of media manipulation and its effects on a democracy (in Summer Semester 2024,
the second author used it to support his course \textit{Media
Manipulation, Propaganda and Fake News} at Coburg
University of Applied Sciences in Germany).

In future work, we aim to support open-source language models \cite{Touvron-etal:2023:ArXiv} to reduce cost and decrease reliance on commercial model vendors. 
We intent to support languages
other than English, and
we plan to expand the tool's capabilities for multi-dimensional content analysis, including hate speech detection, readability scoring, fake news detection/credibility assessment and identifying inappropriate content for children \cite{Fuhr-etal:2018:SIGIRForum}. We also welcome collaborations
with other research teams and
contributions to our effort from the open source
community.


%
%
\bibliographystyle{splncs04}
\bibliography{biasscanner.bib}

\end{document}